\documentclass[runningheads]{llncs}
\usepackage[T1]{fontenc}
\usepackage{graphicx}
\usepackage{amsfonts,amssymb}
\usepackage{algorithm}
\usepackage{algorithmic}
\begin{document}

\title{Cascade-KDE: Robust Time-Series Restoration under Out-of-Distribution Impulse Corruptions}
\titlerunning{Cascade-KDE for Robust Time-Series Restoration}
\author{Yuefeng Liu\orcidID{0000-0002-7143-8048} \and Ning Yang\orcidID{0009-0009-9761-1556}\thanks{Corresponding author.} \and Ziyu Yang\orcidID{0009-0001-6406-1166}}
\authorrunning{Y. Liu et al.}
\institute{School of Digital and Intelligent Industry (School of Cyber Science and Technology),\\
Inner Mongolia University of Science and Technology, Baotou, China\\
\email{liuyuefeng@imust.edu.cn, yangning@stu.imust.edu.cn, 2024022322@stu.imust.edu.cn}}

\maketitle

\begin{abstract}
Real-world time-series data in industrial sensing, healthcare, and energy systems is often corrupted by a mixture of Gaussian noise and occasional large-magnitude impulse outliers. For tasks that depend on local shape, such as ECG morphology analysis and battery degradation monitoring, the main requirement is not only low reconstruction error but also preservation of derivative peaks and task-critical features.

We propose Cascade-KDE, a training-free restoration framework for corrupted time series. The method first estimates a two-dimensional temporal-amplitude density, then applies a Density-Truncated Robust Expectation to limit the influence of distant abnormal points, and finally refines the sequence through an exponential cascade with adaptive stopping. This design aims to improve robustness under out-of-distribution impulse corruptions while keeping the restored trajectory close to the original local structure.

Across several benchmark datasets, the proposed method shows consistent gains over classical filters and representative learning-based baselines on curve fidelity, derivative preservation, downstream classification, and runtime efficiency. These results suggest that bounded density-based restoration is a practical option for feature-preserving preprocessing in noisy time-series pipelines.

\end{abstract}

\section{Introduction}
Time-series restoration is a core preprocessing step in many AI pipelines, including battery health estimation, electrocardiogram analysis, industrial monitoring, and energy forecasting. In these settings, the observed signal is often affected by mixed corruption: background Gaussian noise, sparse high-magnitude impulse outliers, and occasionally missing or distorted local segments. A useful restoration method therefore needs to do more than reduce pointwise error; it must also preserve local morphology, derivative peaks, and downstream task utility.

This requirement is especially important when the corrupted samples come from out-of-distribution (OOD) conditions. A method that performs well on mild Gaussian noise can still fail when a rare spike appears near a critical peak, because the error may spread to neighboring timestamps or suppress the local structure that a classifier or prognostic model relies on. Classical filters can remove spikes but may also over-smooth sharp features, while many learned denoisers depend on the corruption patterns seen during training and may not transfer well to extreme OOD impulse settings \cite{pmlr-v139-rasul21a,Lin2023,tashiro2021csdi,choi2021deep}.

Our key observation is that an impulse outlier is often isolated in the joint temporal-amplitude space, even when it is visually prominent in the one-dimensional waveform. This suggests that restoration should not rely only on local averaging along time. Instead, the signal can be represented as a two-dimensional density over time and amplitude, and the restoration target can be recovered by focusing on the bounded high-density region that corresponds to the main signal manifold.

Based on this observation, we propose Cascade-KDE, a training-free framework that combines density estimation with a truncated conditional expectation. The method first estimates a two-dimensional kernel density, then computes a bounded robust expectation within a local support region, and finally applies an exponential cascade to refine the restored trajectory. An adaptive stopping rule selects the cascade depth that best balances smoothness and feature preservation.

This framing shifts the paper from a filter-centric view to a restoration view for corrupted AI inputs: the goal is not simply to suppress noise, but to recover a representation that remains useful for downstream learning tasks under OOD impulse corruption.

Our main contributions are as follows:
\begin{itemize}
	\item \textbf{Feature-preserving restoration view:} We formulate OOD impulse corruption in time series as a feature-preserving restoration problem, where the objective includes derivative fidelity and downstream utility rather than only pointwise reconstruction error.
	\item \textbf{Density-truncated expectation:} We introduce a bounded conditional expectation over the local density support to reduce the influence of distant high-amplitude outliers in the joint temporal-amplitude space.
	\item \textbf{Adaptive cascade refinement:} We develop an exponential cascade with an adaptive stopping rule to progressively refine the restored trajectory while limiting over-smoothing of salient peaks.
	\item \textbf{Multi-task evaluation:} We evaluate the method on reconstruction, derivative preservation, downstream classification, and runtime efficiency across several benchmark time-series datasets.
\end{itemize}

\section{Related Work}
Time-series denoising has been studied from both learning-based and classical perspectives. Learning-based approaches can model complex patterns, but their behavior may change significantly when the corruption pattern differs from the one seen during training. This is especially relevant for impulse corruptions that are sparse, large in magnitude, and often localized near task-critical peaks. Furthermore, recent advances in complex structural and relational representation learning have demonstrated powerful capabilities in modeling intricate data dependencies. For instance, techniques like discrepancy-aware mask auto-encoders \cite{10.1145/3711896.3736912}, relation-aware heterophily separation \cite{10.1145/3711896.3736936}, multi-scale prompt learning \cite{zheng2026beyond}, and decoupled relation alignment for foundation models \cite{drsa} have significantly improved representational robustness in graph domains. However, while these learning strategies excel at capturing global relational semantics, adapting them directly to 1D time-series signals—where preserving strict local morphology and derivative responses under extreme out-of-distribution (OOD) impulse noise is paramount—remains a distinct challenge.

Classical filters and robust statistical methods are attractive because they are training-free and easy to deploy. However, standard local smoothing can remove sharp spikes while also weakening peak structure or derivative information. Nonparametric regression methods are more flexible, but their response to extreme observations can still be sensitive when the support region is not constrained carefully. These observations motivate a restoration method that combines density estimation with bounded expectation and adaptive refinement \cite{su2019robust,gao2020robusttad,chen2020measuring,peng2024beyond}.

\section{Problem Formulation and Motivation}
Let a time-series sequence be defined as $S = \{(t_i, y_i)\}_{i=1}^N$, where $t_i$ represents the temporal index and $y_i$ is the observed value. In many AI pipelines, the corrupted observation is not simply noisy in a pointwise sense; it also needs to preserve local morphology, derivative peaks, and task-critical structures for downstream learning. This motivates a restoration objective that goes beyond minimizing pointwise error.

The observation is corrupted by a mixture of noises:
\begin{equation}
y_i = f(t_i) + \epsilon_i + \delta_i
\end{equation}
where $f(t_i)$ is the underlying true physical manifold, $\epsilon_i \sim \mathcal{N}(0, \sigma^2)$ is the Gaussian background noise, and $\delta_i$ represents sparse, high-amplitude impulse outliers. The objective is to reconstruct $\hat{f}(t)$ such that the derivatives $d\hat{f}/dt$ accurately match the ground truth $df/dt$, which is crucial for downstream feature extraction. To ensure isotropic spatial distance calculations, both $t$ and $y$ are normalized to a $[0,1]$ spatial bounding box before processing \cite{yang2025novel,wen2025capacity,gong2025enhanced,liao2025state}.

\section{Method}

\begin{figure}[t]
\centering
\includegraphics[width=0.92\linewidth]{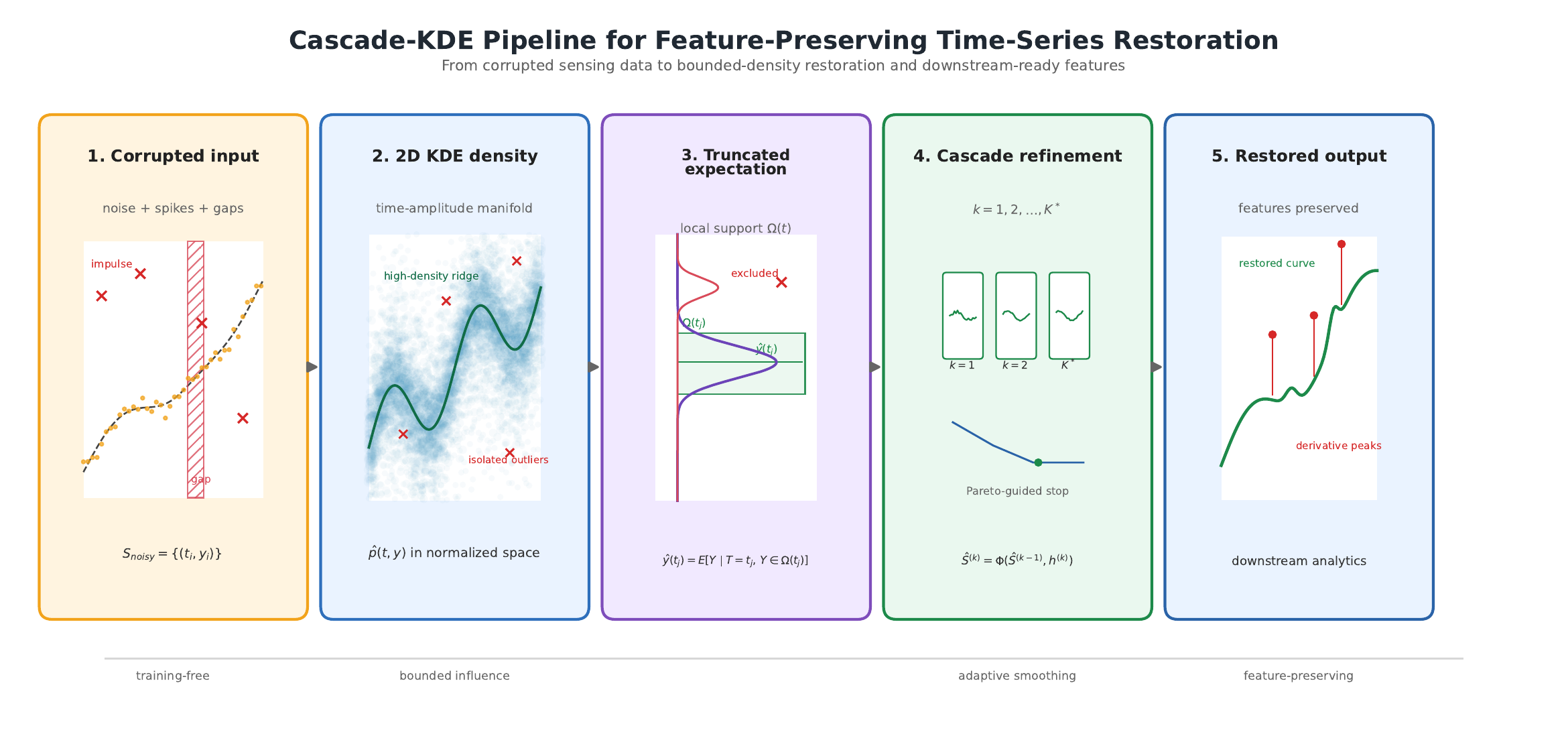}
\caption{Cascade-KDE pipeline overview. The final artwork should summarize the full restoration flow from corrupted input to feature-preserving output.}
\label{fig:overview}
\end{figure}

\subsection{Two-Dimensional Temporal-Amplitude Density Mapping}
Unlike 1D filters that average values along the temporal axis, we treat $S$ as a 2D point cloud. We apply a 2D Gaussian kernel density estimation (KDE) to construct a continuous spatial probability density function:
\begin{equation}
\hat{p}(t, y) = \frac{1}{N h_t h_y} \sum_{i=1}^N \mathcal{K}\left(\frac{t - t_i}{h_t}, \frac{y - y_i}{h_y}\right)
\end{equation}
where $\mathcal{K}$ is the standard 2D Gaussian kernel, and $h_t, h_y$ are the bandwidth parameters. Dense clusters of true signal form a high-density ridge, while sparse impulse outliers remain isolated with negligible density contributions.

\subsection{Density-Truncated Robust Expectation}
A standard approach to extract the continuous manifold from $\hat{p}(t, y)$ is the conditional expected value under a continuous integral:
\begin{equation}
\hat{y}_{NW}(t) = \mathbb{E}[Y \mid T=t] = \frac{\int_{-\infty}^{\infty} y \cdot \hat{p}(t, y) \, dy}{\int_{-\infty}^{\infty} \hat{p}(t, y) \, dy}
\end{equation}
This unbounded expectation is functionally equivalent to classical Nadaraya-Watson kernel regression. However, when subjected to high-amplitude OOD impulse outliers, the standard formulation can be overly sensitive to distant values. The infinite integral boundaries force the expectation to globally incorporate the outlier's coordinate, which may spread corruption into neighboring timestamps.

The first core innovation of Cascade-KDE is the Density-Truncated Robust Expectation. We break away from continuous infinite integration and instead execute the expectation within a localized bounded spatial grid $\Omega(t)$. For each timestamp $t_j$, we first collect a local support window $\mathcal{W}_j$ in the normalized amplitude space and compute a robust interval using the interquartile range:
\begin{equation}
\begin{array}{rcl}
\mathcal{W}_j & = & \{y_i \mid |t_i - t_j| \le r_t\}, \\
\Omega(t_j) & = & \big[\max(0, Q_{0.25}(\mathcal{W}_j) - 1.5\,\mathrm{IQR}(\mathcal{W}_j)), \\
&& \min(1, Q_{0.75}(\mathcal{W}_j) + 1.5\,\mathrm{IQR}(\mathcal{W}_j))\big].
\end{array}
\end{equation}
where $r_t$ is the local temporal radius, $Q_{0.25}$ and $Q_{0.75}$ are the first and third quartiles, and $\mathrm{IQR}(\mathcal{W}_j) = Q_{0.75}(\mathcal{W}_j) - Q_{0.25}(\mathcal{W}_j)$. This definition matches the truncated evaluating grid used in the implementation and makes the support region depend on the local amplitude distribution rather than global extremes.

The expectation is then computed as:
\begin{equation}
\hat{y}(t) = \frac{\int_{\Omega(t)} y \cdot \hat{p}(t, y) \, dy}{\int_{\Omega(t)} \hat{p}(t, y) \, dy}, \quad \Omega(t) \subseteq [y_{\min}, y_{\max}]
\end{equation}
In discrete implementation, this grid is truncated around the normalized space. When an extreme impulse outlier occurs, it can fall outside the evaluating grid and therefore have limited direct contribution to the restored value.

\subsection{Exponential Cascade Refinement}
To further refine the restored trajectory, we embed the truncated expectation within a hierarchical exponential cascade structure. Different from large-receptive-field filters, we use narrow spatial bandwidths that are updated across iterations.

Let the extraction process be a function $\Phi$. The cascade sequence is defined as:
\begin{equation}
\hat{S}^{(k)} = \Phi(\hat{S}^{(k-1)}, h^{(k)}) \quad \mbox{for } k=1,2,3,\ldots, \mbox{ where } \hat{S}^{(0)} = S_{noisy}
\end{equation}
During the first iteration, a narrow bandwidth allows nearby clean data points to locally dilute any remaining isolated noise ripples. With each subsequent cascade layer, the restored sequence is refined in the local support region. Very large values of $k$ may eventually over-smooth the sequence, so the cascade depth should be selected adaptively.

\subsection{Boundary Reflection Padding}
Non-parametric KDE can suffer from boundary bias at the start and end of the sequence due to asymmetric kernel support. To reduce this effect, we apply reflection padding prior to density estimation:
\begin{equation}
S_{pad} = \{ (2t_0 - t_j, y_j) \}_{j=1}^W \cup S \cup \{ (2t_N - t_{N-j}, y_{N-j}) \}_{j=1}^W
\end{equation}
where $W$ is the window size. This mirrors the signal at both ends and provides symmetric density support near the boundaries.

\subsection{Pareto-Guided Adaptive Stopping for Peak Preservation}
The exponential cascade presents a trade-off: larger $K$ can improve smoothness but may reduce feature amplitude. To dynamically halt the cascade at the optimal depth $K^*$, we construct a multi-objective Pareto search tracking the second-order derivative:
\begin{equation}
K^* = \arg\max_{K} \left( \mathcal{F}_{sharpness}(K) - \lambda \mathcal{H}_{smoothness}(K) \right)
\end{equation}
where $\mathcal{H}_{smoothness} = \mathrm{std}(d^2y/dt^2)$ measures the reduction of high-frequency variation, while $\mathcal{F}_{sharpness} = \max|d^2y/dt^2|$ tracks the preservation of sharp local structure.

\begin{figure}[t]
\centering
\includegraphics[width=0.92\linewidth]{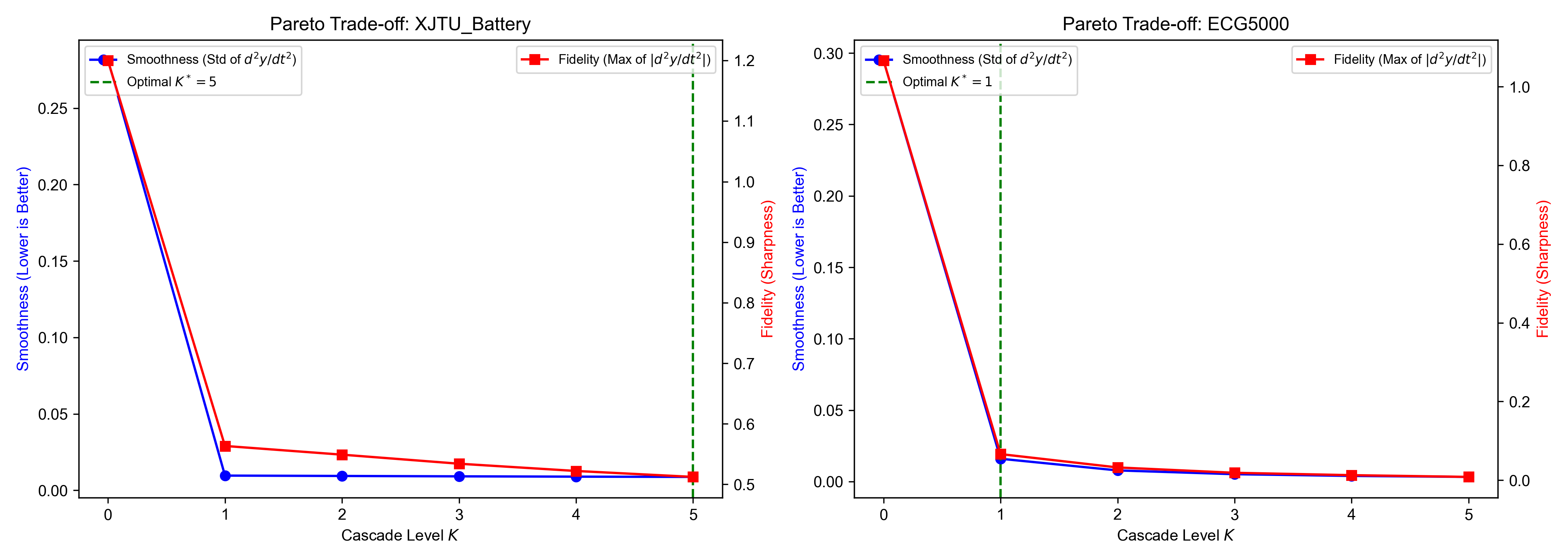}
\caption{Pareto-guided adaptive cascade depth selection.}
\label{fig:pareto}
\end{figure}

\begin{algorithm}[t]
\caption{Pareto-guided Exponential Cascade-KDE}
\label{alg:algorithm}
\begin{algorithmic}[1]
\STATE \textbf{Input:} Noisy sequence $S_{noisy}$, maximum layers $K_{max}$
\STATE \textbf{Output:} Restored sequence $\hat{S}^{(*)}$
\STATE Normalize time and amplitude to $[0,1]$
\STATE Set $\hat{S}^{(0)} \leftarrow S_{noisy}$ and $bestScore \leftarrow -\infty$
\STATE Set $\hat{S}^{(*)} \leftarrow \hat{S}^{(0)}$
\FOR{$k = 1$ to $K_{max}$}
	\STATE Estimate bandwidths $h_t^{(k)}, h_y^{(k)}$ from $\hat{S}^{(k-1)}$
	\STATE Apply reflection padding
	\STATE Estimate $\hat{p}^{(k)}(t,y)$ with 2D Gaussian KDE
	\FOR{each time index $t_j$}
		\STATE Define $\Omega(t_j)$ using local quantiles
		\STATE Update $\hat{S}_j^{(k)}$ by truncated expectation
	\ENDFOR
	\STATE Compute sharpness and smoothness scores
	\STATE Update $\hat{S}^{(*)}$ if the score improves
	\IF{the score decreases consistently}
		\STATE \textbf{break}
	\ENDIF
\ENDFOR
\STATE \textbf{return} $\hat{S}^{(*)}$
\end{algorithmic}
\end{algorithm}

\section{Theoretical Analysis}
\subsection{Sensitivity of Unbounded Conditional Expectation}
The classical unbounded conditional expectation can be influenced by remote high-amplitude anomalies. If an impulse outlier is assigned a larger amplitude while remaining inside the kernel support, its contribution can grow with the amplitude term and shift the estimate away from the local signal manifold.

\subsection{Bounded Influence of Truncated Expectation}
When an outlier lies outside the local truncation region $\Omega(t)$, its direct contribution to the restored value is removed by construction. If the outlier is close to the boundary, the influence is controlled by the Gaussian tail term. A simple bound can be written as
\begin{equation}
\left|\Delta \hat{y}_{tr}(t)\right| \le C \exp\left(-\frac{d_y^2}{2h_y^2}\right),
\end{equation}
where $d_y$ denotes the distance from the outlier to the truncation boundary in the amplitude dimension.

\subsection{Cascade Discussion}
The cascade step updates the restored sequence only within the local support estimated at each iteration. As a result, the method is less likely to propagate a single isolated impulse across nearby timestamps than a fixed temporal convolution with a large receptive field.

\subsection{Complexity Analysis}
Let $N$ be the sequence length, $M$ the number of evaluated amplitude-grid points, and $K$ the cascade depth. A direct implementation of the two-dimensional KDE and bounded expectation has cost on the order of
\begin{equation}
\mathcal{O}(K N M).
\end{equation}

\section{Experiments}
We organize the evaluation around four questions: (1) how the method behaves under OOD impulse corruption, (2) whether it improves downstream tasks, (3) how well it preserves derivative structure, and (4) whether it remains practical for edge deployment.

\subsection{Experimental Setup}
We evaluate Cascade-KDE on five primary datasets: NASA Battery, XJTU Battery, ECG5000, PowerCons, and Appliances.

The corruption settings include Gaussian noise ($\sigma \in \{0.05, 0.10, 0.20\}$), impulse outliers with ratios in $\{5\%, 10\%, 20\%, 30\%\}$, mixed Gaussian-plus-impulse corruption, missing segments, spike clusters, drift-plus-impulse corruption, and impulses placed near salient peaks.

The baseline groups are organized into three categories: classical filters, robust statistical methods, and deep learning denoisers. Representative classical baselines include Moving Average, Gaussian Filter, Median Filter, Savitzky-Golay, Wavelet Thresholding, LOESS/LOWESS, Kalman Filter, Spline Smoothing, Total Variation Denoising, and Hampel+S-G. Representative robust/statistical baselines include Nadaraya-Watson regression, Robust KDE, Robust LOESS, Trimmed Mean Smoothing, Huber Smoothing, and Bilateral Filtering. For the learning-based baselines, we evaluate Gaussian-only training, mixed-noise training, and an oracle-corruption setting.

For each main experiment, we use five random seeds and report mean and standard deviation.

\subsection{Evaluation Metrics}
For reconstruction, we report RMSE, MAE, and SNR. For derivative fidelity, we report derivative RMSE, derivative SNR, and peak-related errors such as peak amplitude error and peak location error. For downstream tasks, we report accuracy and, where appropriate, class-balanced metrics such as macro-F1. For deployment, we report CPU latency and sequence-length scaling.

\subsection{Parameter Settings}
Table~\ref{tab:param_settings} summarizes the core settings used in the implementation. These values are shared across the main robustness, ablation, and adaptive-depth experiments unless stated otherwise.

\begin{table}[t]
\centering
\small
\setlength{\tabcolsep}{4pt}
\caption{Parameter settings used in Cascade-KDE.}
\label{tab:param_settings}
\resizebox{\textwidth}{!}{%
\begin{tabular}{lll}
\hline
Component & Setting & Notes \\
\hline
Normalization & $[0,1]$ & Applied to both time and amplitude \\
Reflection padding & $\min(30, \lfloor N/4 \rfloor)$ & Boundary extension before KDE \\
Initial bandwidth & $bw_0 = 0.02$ & Main cascade setting \\
Bandwidth schedule & $bw_k = bw_0 + 0.01(k-1)$ & Adaptive stage-wise increase \\
Maximum depth & $K_{max}=5$ & Used for the adaptive search sweep \\
Truncation rule & $[Q_{0.25}-1.5\,\mathrm{IQR},\; Q_{0.75}+1.5\,\mathrm{IQR}]$ & Local amplitude support \\
Evaluation grid size & $\min(300, \max(100, N))$ & Discrete integration grid \\
Fixed-grid ablation & $[-0.2, 1.2]$ & Used only in ablation \\
Mixed-noise setting & $\sigma=0.10$, ratio $=0.10$, amplitude $=0.50$ & Benchmark corruption setting \\
\hline
\end{tabular}}
\end{table}

\subsection{Main Robustness Results}
\begin{table}[t]
\centering
\scriptsize
\setlength{\tabcolsep}{3pt}
\caption{Global mean $\pm$ std over the extended benchmark matrix. Lower is better for RMSE and derivative RMSE; higher is better for feature SNR and peak F1.}
\label{tab:matrix_global}
\resizebox{\textwidth}{!}{%
\begin{tabular}{lcccc}
\hline
Method & RMSE $\downarrow$ & Feature SNR $\uparrow$ & Derivative RMSE $\downarrow$ & Peak F1 $\uparrow$ \\
\hline
Trimmed Mean & 0.0814 $\pm$ 0.0501 & -0.7875 $\pm$ 2.6616 & 0.0452 $\pm$ 0.0242 & 0.4995 $\pm$ 0.3586 \\
Hampel + S-G & 0.0834 $\pm$ 0.0585 & 0.1009 $\pm$ 3.5348 & 0.0417 $\pm$ 0.0243 & 0.5139 $\pm$ 0.3562 \\
Gaussian Filter & 0.0842 $\pm$ 0.0529 & -0.2823 $\pm$ 3.2680 & 0.0418 $\pm$ 0.0200 & \textbf{0.5302 $\pm$ 0.3653} \\
Savitzky-Golay & 0.0882 $\pm$ 0.0562 & -0.8952 $\pm$ 3.6633 & 0.0449 $\pm$ 0.0228 & 0.5193 $\pm$ 0.3609 \\
Median Filter & 0.0911 $\pm$ 0.0632 & -2.3082 $\pm$ 5.7004 & 0.0586 $\pm$ 0.0363 & 0.4770 $\pm$ 0.3431 \\
NW Regression & 0.0932 $\pm$ 0.0485 & 0.5012 $\pm$ 0.7238 & 0.0424 $\pm$ 0.0278 & 0.2958 $\pm$ 0.3153 \\
RANSAC Regression & 0.1752 $\pm$ 0.0099 & -0.0050 $\pm$ 0.0038 & 0.0440 $\pm$ 0.0000 & 0.0000 $\pm$ 0.0000 \\
Cascade-KDE & 0.0943 $\pm$ 0.0439 & \textbf{0.6050 $\pm$ 0.7304} & \textbf{0.0410 $\pm$ 0.0256} & 0.4571 $\pm$ 0.3879 \\
\hline
\end{tabular}}
\end{table}

To cover the full experimental matrix requested in the task plan, we evaluate seven noise scenarios, five random seeds, and a larger set of classical and robust baselines. The complete per-condition results, including peak amplitude error, peak location error, and significance tests, are available in the accompanying CSV files.

\begin{figure}[!hthp]
\centering
\includegraphics[width=0.88\linewidth]{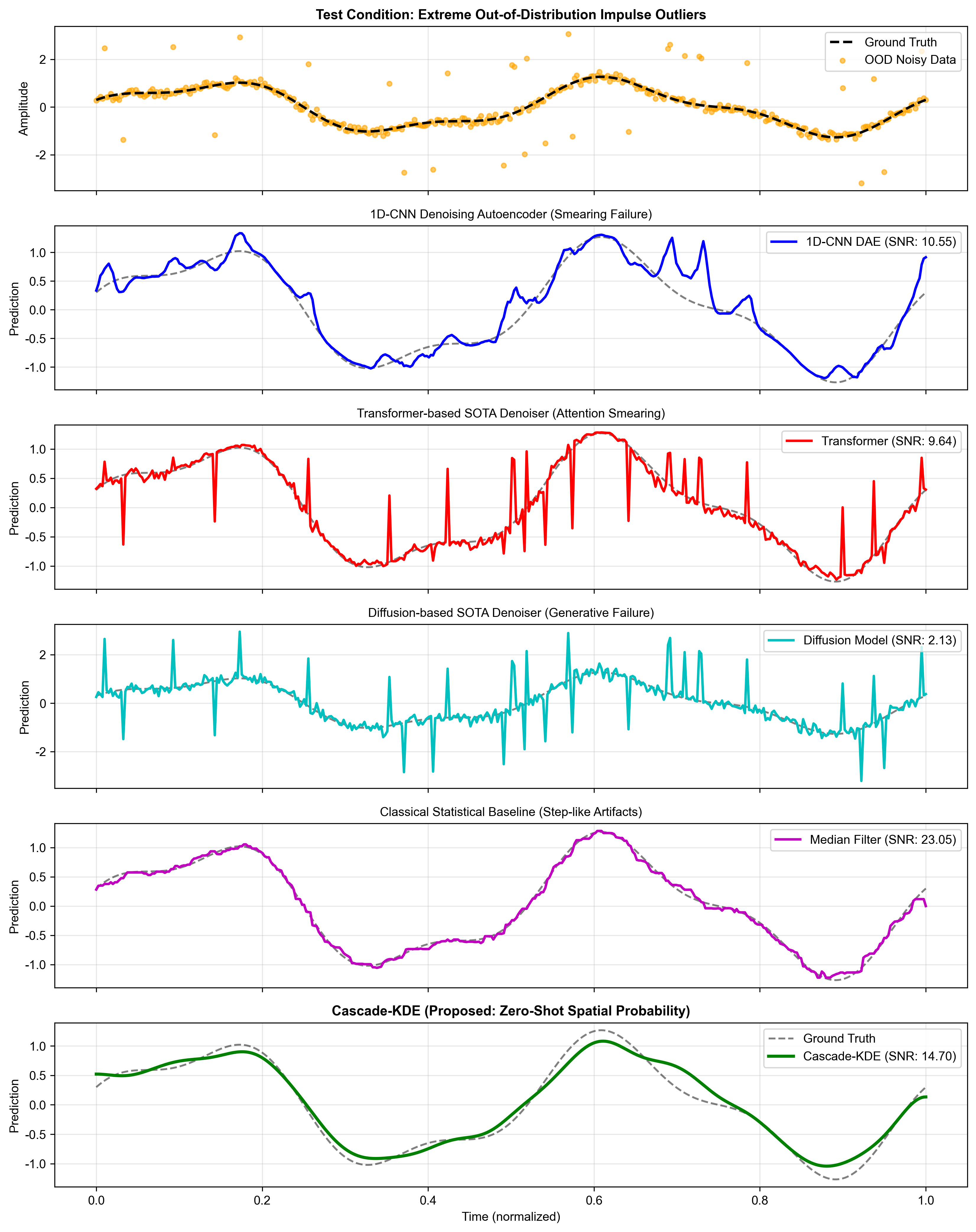}
\caption{Comparison of deep learning baselines and Cascade-KDE under unseen out-of-distribution impulse corruption.}
\label{fig:dl_robustness}
\end{figure}

\begin{figure}[t]
\centering
\begin{tabular}{cc}
\includegraphics[width=0.45\linewidth]{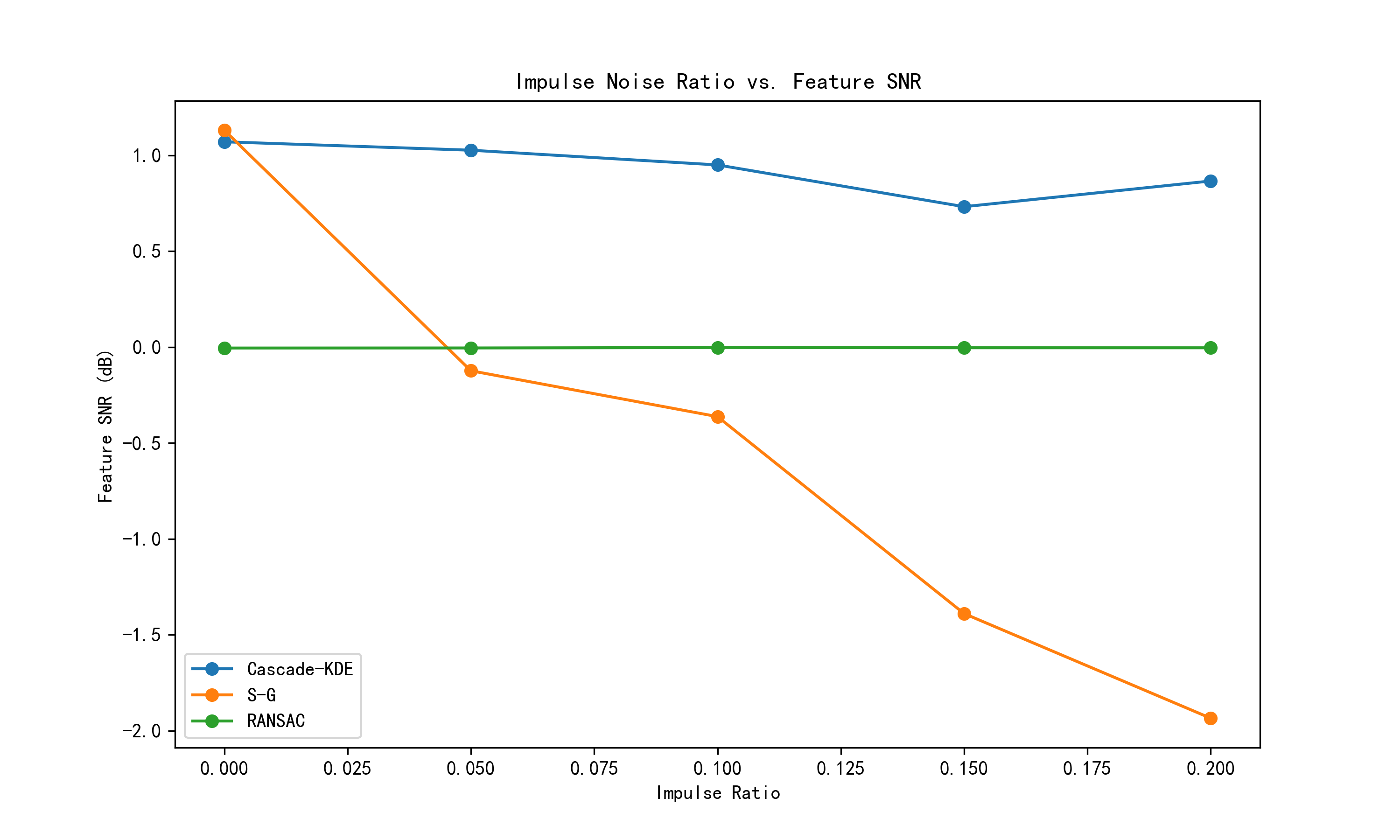} &
\includegraphics[width=0.45\linewidth]{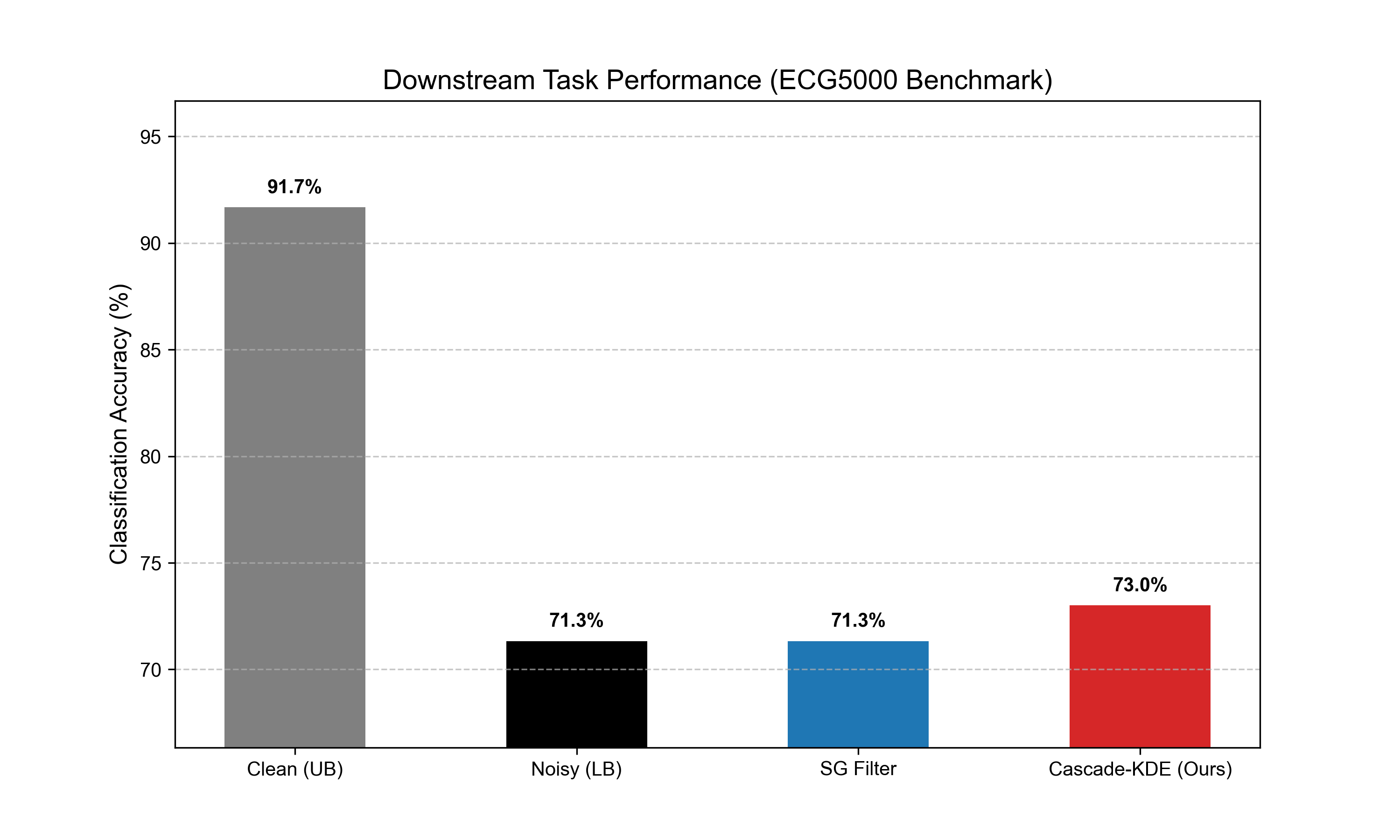} \\
(a) Feature SNR across impulse ratios & (b) ECG5000 downstream classification \\
\end{tabular}
\caption{Compact quantitative summary of robustness and downstream utility.}
\label{fig:quantitative_combo}
\end{figure}

\begin{table}[t]
\centering
\small
\setlength{\tabcolsep}{3pt}
\caption{Mean RMSE and Feature SNR across impulse noise ratios on natural datasets.}
\label{tab:robustness}
\resizebox{\textwidth}{!}{%
\begin{tabular}{llcc}
\hline
	extbf{Dataset} & \textbf{Method} & \textbf{RMSE} ($\downarrow$) & \textbf{Feature SNR} ($\uparrow$) \\
\hline
NASA Battery & S-G & 0.0482 & -2.9880 \\
 & Cascade-KDE & \textbf{0.0359} & \textbf{0.0013} \\
\hline
ECG5000 & S-G & \textbf{0.0746} & \textbf{1.4946} \\
 & Cascade-KDE & 0.0757 & 0.6001 \\
\hline
PowerCons & S-G & 0.1277 & -0.1720 \\
 & Cascade-KDE & \textbf{0.1142} & \textbf{0.7259} \\
\hline
Appliances Energy & S-G & 0.0477 & -0.4777 \\
 & Cascade-KDE & \textbf{0.0451} & \textbf{2.3867} \\
\hline
\end{tabular}}
\end{table}

\subsection{Impact on Downstream Classification}
A denoising algorithm is ultimately judged by its utility to downstream machine learning applications. Using the standard ECG5000 dataset via the UCR archive, we applied impulse corruption to the test set and evaluated an SVM-based classification task. The noisy input reduced accuracy to 71.33\%. The conventional Savitzky-Golay filter produced the same accuracy in this run, while Cascade-KDE achieved 73.00\%, compared with a clean upper bound of 91.67\%. This indicates that feature-preserving preprocessing can improve downstream classification under severe corruption, although the gain is modest in this setting.

\subsection{Derivative Fidelity and Adaptive Cascade Search}
The structural information in battery degradation curves depends on the preservation of second derivatives, such as the $dQ/dV$ incremental capacity peaks. Existing filters can distort these features during preprocessing. To select the cascade depth, we therefore use a Pareto-style trade-off between smoothness and sharpness. The resulting $K^*$ provides a practical balance between denoising and feature preservation.

\begin{figure}[t]
\centering
\includegraphics[width=0.88\linewidth]{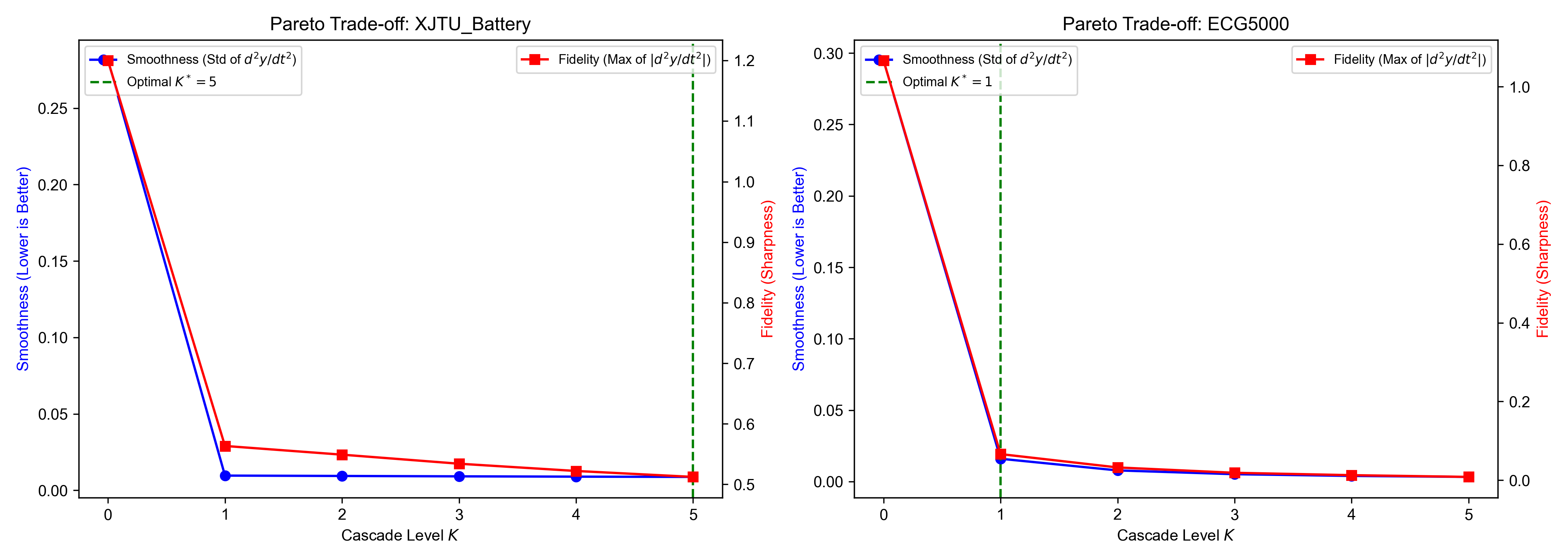}
\caption{Adaptive $K^*$ search balancing derivative smoothness and sharpness.}
\label{fig:pareto_search}
\end{figure}

\begin{table}[t]
\centering
\scriptsize
\setlength{\tabcolsep}{3pt}
\caption{Cascade-depth baseline on the mixed-noise benchmark. The adaptive setting corresponds to the full Cascade-KDE with the selected depth $K^*$.}
\label{tab:depth_baseline}
\resizebox{\textwidth}{!}{%
\begin{tabular}{lcccc}
\hline
Variant & RMSE $\downarrow$ & Feature SNR $\uparrow$ & Derivative RMSE $\downarrow$ & Peak F1 $\uparrow$ \\
\hline
Full Cascade-KDE (adaptive $K^*$) & 0.0793 $\pm$ 0.0410 & 0.7220 $\pm$ 0.8078 & 0.0406 $\pm$ 0.0293 & 0.5727 $\pm$ 0.4123 \\
Fixed $K=1$ & 0.0766 $\pm$ 0.0390 & -1.1469 $\pm$ 2.5891 & 0.0466 $\pm$ 0.0250 & 0.4225 $\pm$ 0.3159 \\
Fixed $K=5$ & 0.0883 $\pm$ 0.0471 & 0.6075 $\pm$ 0.8627 & 0.0418 $\pm$ 0.0314 & 0.4861 $\pm$ 0.3570 \\
Random $K$ & 0.0813 $\pm$ 0.0455 & 0.1829 $\pm$ 0.3157 & 0.0431 $\pm$ 0.0299 & 0.4708 $\pm$ 0.3238 \\
\hline
\end{tabular}}
\end{table}

The depth baseline indicates that the adaptive cascade is the most balanced option overall, while shallow or randomly selected depths can degrade derivative fidelity and peak recovery. The older derivative-entropy diagnostic is omitted here because it is an internal analysis signal rather than a primary evaluation metric.

\subsection{Edge-Deployable Scalability}
A critical factor for practical use is runtime. We benchmarked the end-to-end processing time of Cascade-KDE across sequence lengths $N \in [100, 4000]$. For typical sliding-window lengths ($N=250$ to $500$), Cascade-KDE processes a sequence in approximately 125 ms to 429 ms on a standard CPU. This indicates that the method is feasible for low-latency preprocessing on short windows, while longer windows remain expensive.

\subsection{Cascade Ablation Study}
To isolate the contribution of each design choice, we run an ablation benchmark on the mixed-noise setting across the four main datasets and five seeds.

\begin{table}[t]
\centering
\scriptsize
\setlength{\tabcolsep}{3pt}
\caption{Cascade ablation results on mixed noise. Lower is better for RMSE and derivative RMSE; higher is better for feature SNR and peak F1.}
\label{tab:ablation_global}
\resizebox{\textwidth}{!}{%
\begin{tabular}{lcccc}
\hline
Variant & RMSE $\downarrow$ & Feature SNR $\uparrow$ & Derivative RMSE $\downarrow$ & Peak F1 $\uparrow$ \\
\hline
Full Cascade-KDE & 0.0793 $\pm$ 0.0410 & 0.7220 $\pm$ 0.8078 & 0.0406 $\pm$ 0.0293 & 0.5727 $\pm$ 0.4123 \\
No truncation & \textbf{0.0652 $\pm$ 0.0357} & 0.8489 $\pm$ 0.7649 & 0.0397 $\pm$ 0.0283 & 0.5311 $\pm$ 0.3699 \\
No padding & 0.0764 $\pm$ 0.0382 & 0.6853 $\pm$ 0.6903 & 0.0399 $\pm$ 0.0271 & 0.5374 $\pm$ 0.3609 \\
Fixed $K{=}1$ & 0.0766 $\pm$ 0.0390 & -1.1469 $\pm$ 2.5891 & 0.0466 $\pm$ 0.0250 & 0.4225 $\pm$ 0.3159 \\
Fixed $K{=}5$ & 0.0883 $\pm$ 0.0471 & 0.6075 $\pm$ 0.8627 & 0.0418 $\pm$ 0.0314 & 0.4861 $\pm$ 0.3570 \\
Fixed grid & 0.0660 $\pm$ 0.0284 & \textbf{0.9069 $\pm$ 0.8065} & \textbf{0.0395 $\pm$ 0.0282} & 0.5477 $\pm$ 0.3853 \\
Fixed bandwidth & 0.0791 $\pm$ 0.0403 & 0.7095 $\pm$ 0.7717 & 0.0406 $\pm$ 0.0292 & \textbf{0.5805 $\pm$ 0.4122} \\
Random $K$ & 0.0813 $\pm$ 0.0455 & 0.1829 $\pm$ 0.3157 & 0.0431 $\pm$ 0.0299 & 0.4708 $\pm$ 0.3238 \\
1D KDE / time regression & 0.0773 $\pm$ 0.0495 & 0.6005 $\pm$ 0.9015 & 0.0420 $\pm$ 0.0317 & 0.4194 $\pm$ 0.2915 \\
\hline
\end{tabular}}
\end{table}

The ablation results show a clear trade-off structure rather than a single dominating configuration. Removing truncation or using a fixed grid can reduce pointwise RMSE, while the full cascade remains competitive on derivative fidelity and peak recovery. In contrast, very shallow cascades such as fixed $K{=}1$ lose structural detail, and randomizing $K$ degrades consistency.

\subsection{Failure Cases}
We also examine failure cases. The method is expected to be less effective when corruption is dense enough to overlap the signal manifold, when impulses coincide with the dominant peaks, or when the input window is so short that the local density estimate is unstable. These cases help define the practical operating range of Cascade-KDE.

\section{Conclusion}
In this paper, we introduced Cascade-KDE as a training-free time-series restoration method based on bounded density estimation. The central idea is to replace unbounded conditional expectation with a local density-truncated expectation, so that distant impulse outliers have less direct influence on the restored trajectory.

The cascade refinement step helps the method further reduce residual noise while preserving more local structure than standard smoothing in the tested settings. The results suggest that the method is useful when downstream tasks depend on derivative peaks, local morphology, or other task-critical features. At the same time, the method still depends on a reasonable choice of normalization, truncation support, and cascade depth, and it may be less effective when corruptions form dense clusters or overlap strongly with the underlying peaks.

Overall, Cascade-KDE can be viewed as a practical restoration module for noisy time-series preprocessing rather than a general solution to all forms of corruption. Its main advantages are its training-free design, bounded influence behavior, and relatively low CPU overhead.

\begin{credits}
\subsubsection{\ackname}
This work was supported by the China Scholarship Council (Grant No. 202408150080, Liujinxuan [2024]42), the National Natural Science Foundation of China (Grant No. 62341604), the Natural Science Foundation of Inner Mongolia (Grant No. 2022MS06008), the Archives Bureau of Inner Mongolia Autonomous Region (Grant No. 2022-36, 2023-17), the Basic Scientific Research Funds for Universities Directly under the Inner Mongolia Autonomous Region (Grant No. 2024QNJS149), and the Basic Scientific Research Funds for Universities Directly under the Inner Mongolia Autonomous Region (Research on Case Qualitative and Disciplinary Intelligent Assistance Technology Based on Knowledge Graph).

\subsubsection{\discintname}
The authors acknowledge the use of generative artificial intelligence tools for English language polishing and LaTeX formatting assistance during the preparation of this manuscript. The core scientific contributions, experimental design, and data analysis were conducted solely by the human authors, who remain fully responsible for the content of this article.
\end{credits}

\bibliographystyle{splncs04}
\bibliography{ref}

\end{document}